\pdfoutput=1
\PassOptionsToPackage{table}{xcolor}
\documentclass[11pt]{article}
\usepackage[final]{acl}
\usepackage[table]{xcolor}

\usepackage{times}
\usepackage{multirow}
\usepackage{latexsym}
\usepackage{amssymb}
\usepackage{subcaption}

\usepackage[T1]{fontenc}

\usepackage[utf8]{inputenc}

\usepackage{microtype}

\usepackage{inconsolata}

\usepackage{graphicx}

\title{DenseLoRA: Dense Low-Rank Adaptation of Large Language Models}

\author{
  \textbf{Lin Mu\textsuperscript{1}},
  \textbf{Xiaoyu Wang\textsuperscript{1}},
  \textbf{Li Ni\textsuperscript{1}},
  \textbf{Yang Li\textsuperscript{1}},
  \textbf{Zhize Wu\textsuperscript{2}},
  \\
  \textbf{Peiquan Jin\textsuperscript{3}},
  \textbf{Yiwen Zhang\textsuperscript{1}}\footnotemark[1]
  \\
  \textsuperscript{1}Anhui University,
  \textsuperscript{2}Hefei University,
  \\
  \textsuperscript{3}University of Science and Technology of China,
\\
  \small{
    \texttt{\{mulin, nili, zhangyiwen\}@ahu.edu.cn \{wangxiaoyu,g12114008\}@stu.ahu.edu.cn}
  }
  \\
  \small{
    \texttt{wuzz@hfuu.edu.cn jpq@ustc.edu.cn}
  }
}

\begin{document}
\maketitle

\begin{abstract}
Low-rank adaptation (LoRA) has been developed as an efficient approach for adapting large language models (LLMs) by fine-tuning two low-rank matrices, thereby reducing the number of trainable parameters. However, prior research indicates that many of the weights in these matrices are redundant, leading to inefficiencies in parameter utilization. To address this limitation, we introduce \textbf{Dense} \textbf{Lo}w-\textbf{R}ank \textbf{A}daptation (\textbf{DenseLoRA}), a novel approach that enhances parameter efficiency while achieving superior performance compared to LoRA.
DenseLoRA builds upon the concept of representation fine-tuning, incorporating a single $Encoder$-$Decoder$ to refine and compress hidden representations across all adaptation layers before applying adaptation. Instead of relying on two redundant low-rank matrices as in LoRA, DenseLoRA adapts LLMs through a dense low-rank matrix, improving parameter utilization and adaptation efficiency. We evaluate DenseLoRA on various benchmarks, showing that it achieves 83.8\% accuracy with only 0.01\% of trainable parameters, compared to LoRA's 80.8\% accuracy with 0.70\% of trainable parameters on LLaMA3-8B. Additionally, we conduct extensive experiments to systematically assess the impact of DenseLoRA’s components on overall model performance. Code is available at \href{https://github.com/mulin-ahu/DenseLoRA}{https://github.com/mulin-ahu/DenseLoRA}.

\end{abstract}

\renewcommand{\thefootnote}{\fnsymbol{footnote}}
\footnotetext[1]{Corresponding author}

\section{Introduction}
Large language models (LLMs)~\cite{gpt3, touvron2023llama}, pre-trained on vast general-domain datasets, have shown remarkable generalization capabilities across a diverse range of downstream tasks~\cite{llmcvpr,large2023}. A common approach for adapting LLMs to new tasks is full fine-tuning, which involves retraining all model parameters. However, as LLMs continue to scale, fully fine-tuning all parameters becomes increasingly impractical due to escalating computational and memory costs, especially in resource-constrained settings.

\begin{figure}[t] 
	\centering
	\includegraphics[width=\linewidth]{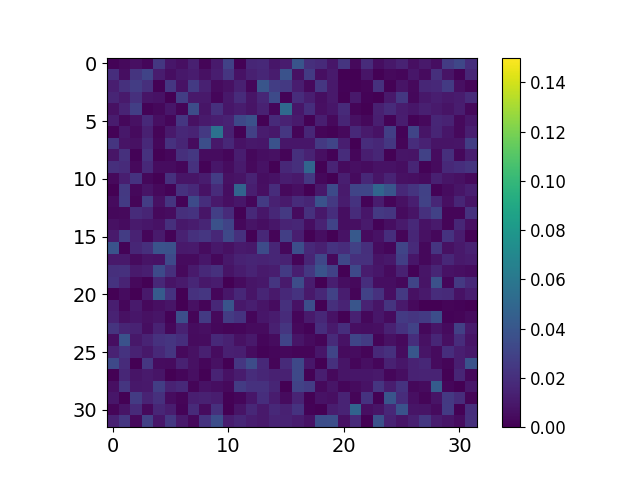}
  \caption{Increments of a low-rank matrix of LoRA during training. We randomly select the slices of a small set of low-rank matrix for demonstration.}
  \label{fig:motivation}
\end{figure}

To address this challenge, researchers have explored parameter-efficient fine-tuning (PEFT)~\cite{adapter, peftsurvey2023}, a class of methods that adapt LLMs by fine-tuning only a small subset of task-specific parameters while keeping the rest of the model frozen. PEFT achieves performance comparable to full fine-tuning while significantly reducing the trainable parameters. Among these methods, low-rank adaptation (LoRA)~\cite{hu2022lora} has emerged as a particularly effective technique, as it maintains the architecture of the LLMs and preserves inference efficiency. LoRA draws on the assumption that the adaptation weights of LLMs exhibit a low "intrinsic rank," allowing updates to be approximated efficiently with low-rank matrices. Despite its efficiency, LoRA has limitations. Empirical studies reveal that a significant portion of the weights within LoRA’s low-rank matrices remain inactive during adaptation, leading to redundancy. As shown in Figure~\ref{fig:motivation}, many parameters in these matrices exhibit near-zero increments, indicating that these weights do not contribute meaningfully to the adaptation. While recent LoRA variants~\cite{zhang2023adaptive, roselora, melora, yin2024lofit} attempt to address these inefficiencies by selectively identifying impactful weights, they remain constrained of traditional low-rank adaptation framework. This raises a fundamental research question:

\textit{"Can we develop a low-rank adaptation method that enhances model performance while leveraging a denser structure to achieve greater efficiency with fewer trainable parameters?"}

To answer this, we propose a paradigm shift: rather than relying solely on modifying weight matrices, we explore refining the hidden representations themselves. Drawing inspiration from representation fine-tuning~\cite{wu2024reft, red} techniques, we aim to enhance expressivity while maintaining efficiency. 
Building on this insight, we introduce \textbf{Dense} \textbf{Lo}w-\textbf{R}ank \textbf{A}daptation (\textbf{DenseLoRA}), a novel framework that integrates low-rank adaptation with representation fine-tuning to improve the adaptation efficiency of LLMs. Specifically, DenseLoRA refines and compresses hidden representations before adaptation, allowing for a more efficient adaptation process with a dense low-rank matrix. 

In DenseLoRA, a $Encoder$ module first refines and compresses the hidden representations across all adaptation layers, preserving essential task-relevant information. A dense low-rank matrix then adapts the compressed representations at each layer. Finally, a $Decoder$ module reconstructs the refined representations, ensuring seamless integration with the pre-trained model. Notably, the $Encoder$ and $Decoder$ are shared across all adaptation  layers, which enhances efficiency and reduces redundancy. Unlike LoRA, which relies on two redundant matrices, DenseLoRA leverages a dense and small matrix to achieve a more efficient and compact adaptation process. This results in a significant reduction in trainable parameters while maintaining effective low-rank adaptation of the pre-trained weight matrix $W_0$.

Our main contributions can be summarized as follows:

\begin{itemize}
\item We introduce DenseLoRA, a novel PEFT method that enhances low-rank adaptation by utilizing a dense and small matrix. This approach leads to more efficient parameter updates, reducing redundancy. 
\item We integrate low-rank adaptation with representation fine-tuning, enabling DenseLoRA to enhance the expressivity of the model while maintaining computational efficiency.
\item We conduct extensive experiments to evaluate DenseLoRA performance on various tasks. Notably, DenseLoRA adapts only 0.01\% of trainable parameters while achieving 83.8\% accuracy, surpassing LoRA's 80.8\% accuracy with 0.70\% trainable parameters on commonsense reasoning tasks. Furthermore, we provide an in-depth analysis of its components and their impact on performance.
\end{itemize}

\begin{figure*}[t] %
    \begin{minipage}[t]{0.5\linewidth}
		\centering
		\includegraphics[width=0.95\linewidth]{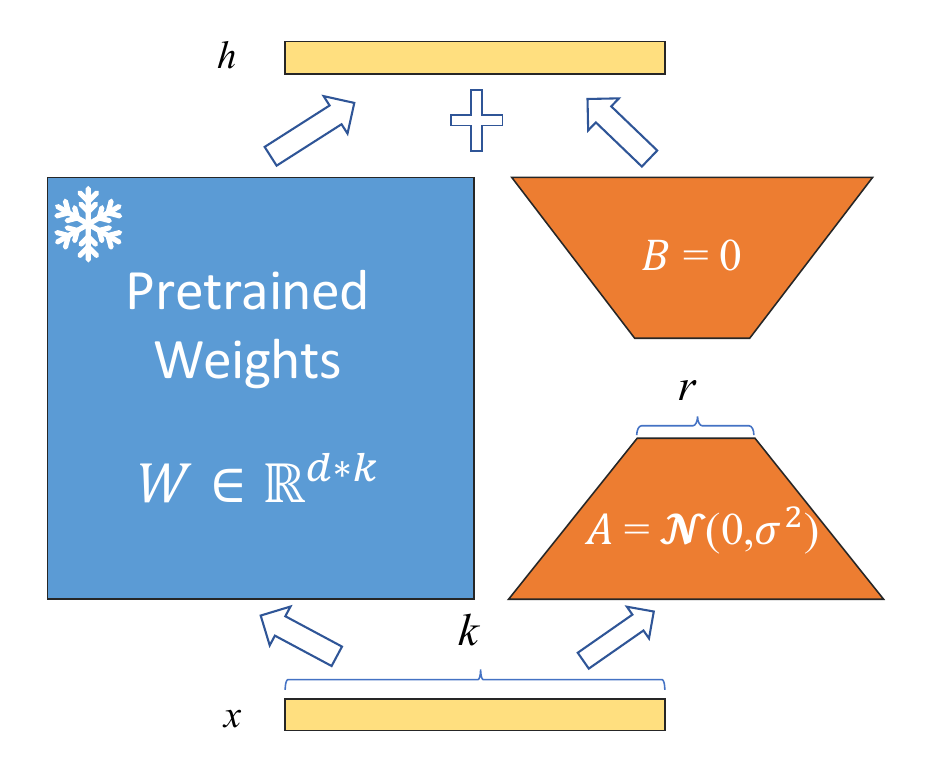}
		\subcaption{(a) LoRA}
        \label{fig:lora}
    \end{minipage}
    \begin{minipage}[t]{0.5\linewidth}
		\centering
		\includegraphics[width=0.95\linewidth]{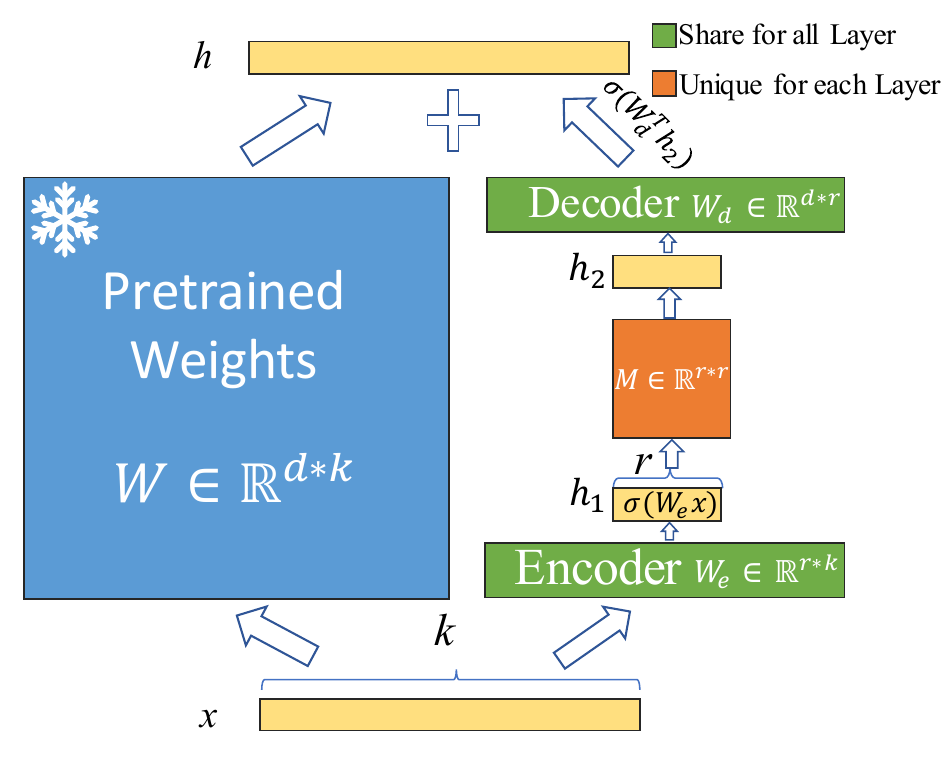}
		\subcaption{\\ (b) DenseLoRA}
        \label{fig:multiple}
	\end{minipage}
  \caption{Framework comparison of LoRA(left) and DenseLoRA(right).}
  \label{fig:architecture}
\end{figure*}

\section{Background}
\subsection{Low-Rank Adaptation (LoRA)}
The core hypothesis of LoRA~\cite{hu2022lora} is that during the fine-tuning, the weight updates exhibit a low "intrinsic rank". Building on this observation, LoRA freezes the pre-trained weights of LLMs and incrementally updates these weights by utilizing the product of two trainable low-rank matrices. This approach has been demonstrated to achieve performance comparable to full fine-tuning across numerous benchmarks while significantly reducing training parameters.
To formally describe LoRA’s adaptation process, let $W_0 \in \mathbb{R}^ {d\times k}$ denote the pre-trained weight matrix. Instead of directly updating $W_0$, LoRA applies incremental updates using two low-rank matrices, expressed as $ \Delta W = BA $, where $B \in \mathbb{R}^{d\times r}$, $A \in \mathbb{R}^{r\times k}$. The rank $r$ is significantly smaller than both $d$ and $k$ (i.e. $r << min(d,k)$). This approach substantially reduces the number of trainable parameters compared to full fine-tuning. By incorporating these incremental updates, the adapted hidden representation $\hat{h}\in \mathbb{R}^{d}$ represented as:
\begin{equation}\label{equ:lora}
    \hat{h} = (W_0 + \Delta W)h = W_0h+ BAh
\end{equation}
Here, $h \in \mathbb{R}^{k}$ indicates the hidden representation before adaptation. $W_0$ remains frozen during fine-tuning. To ensure stability during fine-tuning, matrix $A$ is initialized using a uniform Kamining distribution~\cite{kaiming}, while $B$ is set to zero, ensuring that $\Delta W=0$ at the beginning of fine-tuning. As shown in Eq.(\ref{equ:lora}), during inference, LoRA can merge the $W_0$ and $\Delta W$ (i.e. $W^{'}=W_0 + \Delta W$). This property ensures that LoRA does not introduce any additional inference latency compared to the original model.

\subsection{Representation Fine-tuning}

Recently, several studies have explored representation fine-tuning~\citep{wu2024reft, red} technique. Instead of adapting model weights, these methods focus on refining hidden representations directly, enabling task-specific adjustments without modifying weight matrices. For example, Red~\cite{red} refined hidden representations by introducing two learnable components: scaling vector $l_{scaling} \in \mathbb{R}^{d}$ and bias vector $l_{bias}\in \mathbb{R}^{d}$. This process can be mathematically expressed as follows:
\begin{equation}\label{equ:reft}
    \hat{h} = l_{scaling} \odot h + l_{bias}
\end{equation}
Where $\odot$ represents Hadamard product, $h \in \mathbb{R}^{d}$ indicates the hidden representation.

\section{Methodology}
To address the inefficiencies of existing low-rank adaptation methods, we propose Dense Low-Rank Adaptation (DenseLoRA), a novel framework that integrates low-rank adaptation with representation fine-tuning. DenseLoRA overcomes the redundancy issues observed in vanilla LoRA approaches by adapting a dense, small low-rank matrix, ensuring a more efficient use of parameters. Specifically, DenseLoRA introduces a structured three-stage process: (1) An $Encoder$ refines and compresses hidden representations; (2) A denser low-rank adaptation module adapts the model; and (3) A $Decoder$ reconstructs the refined representations to ensure seamless integration with the pre-trained model. This structured approach distinguishes DenseLoRA from vanilla LoRA, which relies on two redundancy low-rank matrices. 

\subsection{DenseLoRA Architecture}
To realize this three-stage adaptation process, DenseLoRA introduces a novel architecture (illustrated in Figure~\ref{fig:architecture}) that integrates an $Encoder$-$Decoder$ mechanism with low-rank adaptation. This architecture refines and compresses hidden representations before adaptation, reducing redundancy and significantly lowering the number of trainable parameters while maintaining the model expressively. At the core of this architecture is the $Encoder$-$Decoder$ mechanism, implemented as fully connected neural networks, which refines representations and adapts LLMs through the following three stages: 
\begin{itemize}
    \item \textbf{Compression:} The $Encoder$ applies a transformation using weights $W_e \in \mathbb{R}^{r \times k}$ to refine and compress the hidden representation $h \in \mathbb{R}^k$ into a lower-dimensional representation. This is followed by an activation function $\sigma(\cdot)$, producing a compressed representation $h^{'} \in \mathbb{R}^r$. 
    \item \textbf{Adaptation:} To fine-tune the model for downstream tasks, a dense low-rank matrix $\textbf{\textit{M}} \in \mathbb{R}^{r\times r}$ is applied to the compressed representation. This step adapts the model while keeping the pre-trained weight matrix $W_0$ frozen.
    \item \textbf{Reconstruction:} The $Decoder$ then reconstructs the adapted representation back to the original hidden dimension using weights $W_d \in \mathbb{R}^{d \times r}$, followed by an activation function. This reconstruction step ensures that the adapted representation seamlessly integrates with the frozen pre-trained model, preserving expressivity while maintaining efficiency. 
\end{itemize}
The overall adaptation process in DenseLoRA can be mathematically formulated as follows, combining the frozen pre-trained weights $W_0$ with the refined and adapted hidden representations:
\begin{equation}\label{equ:denselora}
    \hat{h} = W_0 h + Decoder(\textbf{\textit{M}} Encoder(h))
\end{equation}

To enhance parameter efficiency, DenseLoRA shares a single $Encoder$-$Decoder$ across all adaptation layers. This strategy reduces redundancy and significantly lowers the number of trainable parameters. To maintain layer-specific adaptability, DenseLoRA fine-tunes unique low-rank matrices $\textbf{\textit{M}} \in \mathbb{R}^{r \times r}$ for each adaptation layer, ensuring that adaptation remains flexible and effective across different layers. The transformations applied by the $Encoder$ and $Decoder$ can be mathematically expressed as follows, capturing how hidden representations are refined and reconstructed:
\begin{equation}\label{equ:encoder}
    h^{'} = Encoder(h) = \sigma(W_e h)
\end{equation}
\begin{equation}\label{equ:decoder}
    \hat{h} = Decoder(h^{'}) = \sigma(W_d^T h^{'})
\end{equation}

\subsection{Parameter Analysis}
\textbf{Initialization Strategies:}
DenseLoRA employs two categories of matrices: Shared Matrices and Unique Matrices, each initialized to ensure stability and efficiency during training. Shared Matrices: These include $W_{e} \in \mathbb{R}^{r \times k}$ and $W_{d} \in \mathbb{R}^{d \times r}$, which are shared across different adaptation layers. The $W_{e}$ matrix is initialized using Kaiming initialization~\cite{kaiming}, while the $W_{d}$ matrix is initialized to zeros. This ensures that the $W_{d}$ matrix does not interfere with the output during the first forward pass. Unique Matrices: These include $\textbf{\textit{M}} \in \mathbb{R}^{r \times r}$, which is unique to each adaptation layer. Like the matrices $W_{e}$, $\textbf{\textit{M}}$ is also initialized using Kaiming initialization~\cite{kaiming} to promote efficient and stable training.

By combining shared matrices for parameter efficiency and unique matrices for layer-specific adaptability, DenseLoRA achieves an effective balance between fine-tuning flexibility and computational cost.

\begin{table*}[t]
    
    \begin{center} 
    \renewcommand{\arraystretch}{1.1}
    \setlength{\tabcolsep}{3pt}
    \resizebox{\linewidth}{!}{
    \begin{tabular}{llcccccccccc}
        \hline 
        \textbf{LLM} & \textbf{Method} & \textbf{ Param(\%)} & \textbf{BoolQ} & \textbf{PIQA} &  \textbf{SIQA} & \textbf{HellaS.} & \textbf{WinoG.} & \textbf{ARC-e} & \textbf{ARC-c} &  \textbf{OBQA} & \textbf{Avg.}\\ \hline
        ChatGPT & - & - & 73.1 & 85.4 & 68.5  & 78.5& 66.1 & 89.8 & 79.9 &  74.8 & 77.0\\
       \hline
        \cellcolor{white}\multirow{4}*{LLaMA2-7B} 
        & LoRA   & 0.83 & 69.8 & 79.9 & 79.5 & 83.6 & 82.6 & 79.8 & 64.7 & 81.0 & 77.6 \\
        & \cellcolor{gray!20}DenseLoRA$^{\ast}$ & \cellcolor{gray!20} \textbf{0.01}& \cellcolor{gray!20}69.9 & \cellcolor{gray!20}79.5& \cellcolor{gray!20}78.2& \cellcolor{gray!20}83.0&\cellcolor{gray!20}78.0& \cellcolor{gray!20}81.5& \cellcolor{gray!20}63.8 & \cellcolor{gray!20}76.6& \cellcolor{gray!20}76.3\\
        
        & \cellcolor{gray!20}DenseLoRA$^{\ast\ast}$ & \cellcolor{gray!20}\textbf{0.03}& \cellcolor{gray!20}71.3& \cellcolor{gray!20}81.0& \cellcolor{gray!20}78.9& \cellcolor{gray!20}85.0&\cellcolor{gray!20}79.5& \cellcolor{gray!20}82.4& \cellcolor{gray!20}65.4 &\cellcolor{gray!20}76.2 &\cellcolor{gray!20}77.5 \\
        & \cellcolor{gray!20}DenseLoRA$^{\ast\ast\ast}$ & \cellcolor{gray!20}\textbf{0.06}& \cellcolor{gray!20}70.2& \cellcolor{gray!20}81.8& \cellcolor{gray!20}78.8&\cellcolor{gray!20}90.0 & \cellcolor{gray!20}81.9& \cellcolor{gray!20}66.2& \cellcolor{gray!20}82.6& \cellcolor{gray!20}79.2&\cellcolor{gray!20}78.8 \\ \hline
        \multirow{10}*{LLaMA3-8B} 
        & LoKr    & 0.01 & 65.1 & 81.6 & 78.7 & 92.0 & 82.1 & 89.2 & 76.7 & 80.9 & 80.9 \\
        & NoRA    & 0.10 & 73.3 & 86.4 & 79.1 & 94.1 & 84.3 & 88.2 & 77.5 & 85.0 & 83.1 \\
        & VeRA$^{\ddagger}$ & 0.01 & 62.2& 81.6& 64.8& 54.5&6.18 & 84.4& 67.2 & 64.6 & 67.7\\
        & AdaLoRA & 0.35 & 75.1 & 86.4 & 76.7 & 75.4 & 83.3 & 90.4 & 79.1 & 81.4 & 81.4 \\
        & LoRA    & 0.70 & 70.8 & 85.2 & 79.9 & 91.7 & 84.3 & 84.2 & 71.2 & 79.0 & 80.8 \\
        & DoRA$^{\dagger}$ & 0.35 & 74.5 & 88.8 & 80.3 & 95.5 & 84.7 & 90.1 & 79.1 & 87.2 & 85.0\\
        & DoRA    & 0.71 & 74.6 & 89.3 & 79.9 & 95.5 & 85.6 & 90.5 & 80.4 & 85.8 & 85.2 \\
        & \cellcolor{gray!20}DenseLoRA$^{\ast}$ & \cellcolor{gray!20}\textbf{0.01}& \cellcolor{gray!20}72.3& \cellcolor{gray!20}87.5&	\cellcolor{gray!20}79.8&\cellcolor{gray!20}93.5&\cellcolor{gray!20}85.2&\cellcolor{gray!20}89.8&	\cellcolor{gray!20}78.2&	\cellcolor{gray!20}84.0 & \cellcolor{gray!20}83.8\\ 
        & \cellcolor{gray!20}DenseLoRA$^{\ast\ast}$ & \cellcolor{gray!20}\textbf{0.02}& \cellcolor{gray!20}74.3& \cellcolor{gray!20}88.0	&\cellcolor{gray!20}80.3	&\cellcolor{gray!20}94.5	&\cellcolor{gray!20}86.0	&\cellcolor{gray!20}89.7&\cellcolor{gray!20}78.7	&\cellcolor{gray!20}85.6&\cellcolor{gray!20}84.6\\
        
        & \cellcolor{gray!20}DenseLoRA$^{\ast\ast\ast}$ & \cellcolor{gray!20}\textbf{0.06}& \cellcolor{gray!20}74.1& \cellcolor{gray!20}88.9 & \cellcolor{gray!20}80.3 &\cellcolor{gray!20}95.0 & \cellcolor{gray!20}87.0	&\cellcolor{gray!20}90.0	&\cellcolor{gray!20}79.2	&\cellcolor{gray!20}85.6 & \cellcolor{gray!20}85.0\\
        \hline 
    \end{tabular}
    }
    \end{center}
    \caption{Accuracy(\%) comparison of various methods fine-tuning LLaMA2-7B and LLaMA3-8B on the commonsense reasoning tasks. $^\dagger$ denotes that rank is equal to 32 in DoRA~\cite{liu2024dora}. $^\ddagger$ denotes that we experiment on the commonsense reasoning task using the same settings in VeRA~\cite{liu2024dora}, with a rank equal to 256. $^\ast$ denotes $r=16$. $^{\ast\ast}$ denotes $r=32$. $^{\ast\ast\ast}$ denotes $r=64$.}
    \label{tab:mainresults}
\end{table*}

\textbf{Parameter Count:}
To effectively fine-tune large language models (LLMs) while maintaining parameter efficiency, DenseLoRA significantly reduces the number of trainable parameters compared to existing LoRA variants. Let $l$ represent the number of adaptation layers in LLMs, and $k$ and $d$ represent the input and output dimensions, respectively. The parameter count for different methods is summarized as follows:
\begin{itemize}
    \item \textbf{Full Fine-Tuning (FFT)}: The total number of trainable parameters is $|\Theta| = l \times d \times k$.
    \item \textbf{LoRA}: LoRA reduces the parameter count to $|\Theta| = l \times (d+ k)\times r$, here rank $r<<min(d, k)$.
    \item \textbf{DenseLoRA}: Unlike LoRA, which relies on two redundant low-rank matrices, DenseLoRA optimizes efficiency by leveraging a denser structure, resulting in the following parameter count: $|\Theta| = (d + k + l \times r) \times r$, here rank $r<<min(d, k)$.
\end{itemize}

To illustrate this advantage, we compare trainable parameters in real-world LLM settings:
\begin{itemize}
    \item In LLaMA2-7B with $r=16$, LoRA require approximate $28M$ trainable parameters, while DenseLoRA reduces this to $0.9M$, achieving a $30\times$ reduction
    \item In LLaMA3-8B with $r=16$, DenseLoRA maintains a similar reduction relative to LoRA, confirming its scalability across models.
\end{itemize}
These results demonstrate that DenseLoRA has significantly fewer trainable parameters, making it a highly scalable and computationally efficient solution for fine-tuning LLMs.

\section{Experiments}
In this section, we conduct extensive experiments to evaluate the effectiveness of DenseLoRA across various tasks. Our evaluation follows a structured approach to ensure a comprehensive analysis.

First, we compare DenseLoRA with LoRA and its variants by fine-tuning LLaMA2-7B and LLaMA3-8B on commonsense reasoning tasks. Next, we extend our analysis to arithmetic reasoning, focusing on fine-tuning LLaMA3-8B models. To assess the performance of DenseLoRA under limited data availability, we conduct experiments by sampling a subset of the original training data and analyzing its impact on model performance. Additionally, we investigate the optimal tuning granularity by analyzing which weight matrices within the transformer architecture benefit the most from adaptation using DenseLoRA. Finally, we delve deeper into the mechanics of DenseLoRA by investigating the effects of fine-tuning the $Encoder$ and $Decoder$ components and comparing the adaptation matrix $\textbf{\textit{M}}$ to $A$ and $B$ used in LoRA.

\subsection{Commonsense Reasoning}
To evaluate DenseLoRA’s performance on commonsense reasoning tasks, we fine-tune LLaMA2-7B and LLaMA3-8B using a dataset comprising 170k training samples~\cite{hu-etal-2023-llm}. These samples are drawn from the training sets of eight commonsense reasoning benchmarks (see Appendix~\ref{sec:dataset} for details):
1) BoolQ~\cite{clark-etal-2019-boolq}; 
2) PIQA~\cite{bisk2020piqa}; 
3) SIQA~\cite{sap-etal-2019-social}; 
4) HellaS. (HellaSwag)~\cite{zellers-etal-2019-hellaswag}; 
5) WinoG. (WinoGrande)~\cite{WinoGrande}; 
6) ARC-c and ARC-e~\cite{ARC}; 
7)OBQA~\cite{openbook}. 
All the experiments are conducted using 4 Nvidia 24GB 3090 GPUs, with the training hyperparameters detailed in Appendix~\ref{sec:Hyperparameters}. 

For comparative analysis, we evaluate DenseLoRA against several LoRA variant methods:
1) ChatGPT~\cite{cot}, which applies a zero-shot chain of thought approach~\cite{zero-shot-cot} on GPT-3.5-turbo;
2) LoRA~\cite{hu2022lora}, which updates a weight matrix via a low-rank matrices $AB$;
3) LoKr~\cite{lokr}, which utilizes Kronecker products for weight matrix decomposition, significantly reducing trainable parameters;
4) NoRA~\citep{NoRA}, which adopts a dual-layer nested structure with singular value decomposition (SVD), effectively leveraging original matrix knowledge while reducing tunable parameters.
5) VeRA~\cite{kopiczko2024vera}, which employs a single pair of shared low-rank matrices across all layers and fine-tunes small scaling vectors;
6) AdaLoRA~\cite{zhang2023adaptive}, which fine-tunes pre-trained weights using SVD;
7) DoRA~\cite{liu2024dora}, which decomposes the pre-trained weight into magnitude and directional components for fine-tuning.

\begin{table*}[t]
    \begin{center} 
    \begin{tabular}{lccccccc}
    \hline 
    \textbf{Method}&\textbf{Params(\%)}& \textbf{GSM8K} &\textbf{AQUA}& \textbf{AddSub} & \textbf{SVAMP} & \textbf{Avg.} \\ \hline
     LoRA & 0.70 & 47.1 &18.1 &  90.6 & 71.9  &   56.9  \\
     DenseLoRA$^{\ast}$  & 0.02 & 45.5 & \textbf{20.5}  & 73.5  &  92.1& 57.5  \\
    DenseLoRA$^{\ast\ast}$ &0.06 & \textbf{47.2} & 19.7 & \textbf{92.4}& \textbf{74.5}   & \textbf{58.5}      \\
    \hline
    \end{tabular}
    \end{center}
    \caption{Accuracy(\%) comparison of DenseLoRA and LoRA fine-tuning LLaMA3-8B on four arithmetic reasoning tasks. $^{\ast}$ denotes $r=32$. $^{\ast\ast}$ denotes $r=64$}
    \label{tab:mathresult}
\end{table*}

\textbf{Main results.} Table~\ref{tab:mainresults} presents the results of commonsense reasoning tasks on LLaMA2-7B and LLaMA3-8B across different ranks (16, 32, and 64). Key findings are:

1) \textbf{Excellent Performance}: DenseLoRA achieves an average accuracy of 85\% on LLaMA3-8B, outperforming LoRA (80.8\%) by 4.1\%, while requiring only 10\% of the trainable parameters used by LoRA. Notably, when fine-tunes just 0.01\% of the parameters, DenseLoRA still maintains 83.8\% accuracy, outperforming LoRA while reducing trainable parameters by a factor of 70. This demonstrates DenseLoRA’s ability to achieve high accuracy with minimal adaptation overhead.

2) \textbf{Parameter Efficiency}: Compared to other parameter-efficient LoRA variants (VeRA, LoKr, and NoRA), DenseLoRA achieves higher accuracy. For instance, it achieves an accuracy of 83.8\%, outperforming LoKr by 2.9\%. While LoKr reduces training parameters via Kronecker products, it requires more computational resources (training time and GPU memory) than LoRA~\cite{wu2024mixtureofsubspaceslowrankadaptation}. In contrast, DenseLoRA maintains similar computational costs to LoRA while offering superior performance, making it a more practical solution.

3) \textbf{Rank Robustness}: DenseLoRA demonstrates strong rank robustness, maintaining high accuracy across rank configurations (16, 32, 64). As the rank increases, its average accuracy consistently improves. For example, at rank 16, DenseLoRA fine-tunes only 1/70 of the trainable parameters used by LoRA, yet outperforms LoRA. At rank 64, DenseLoRA's trainable parameter increases to 0.06\%, achieving 85\% accuracy, surpassing LoRA (80.8\%) and approaching DoRA (85.2\%). 

\subsection{Arithmetic Reasoning}
To evaluate the effectiveness of DenseLoRA on the arithmetic reasoning task, we fine-tuned LLaMA3-8B on the Math10K~\cite{hu-etal-2023-llm} dataset and evaluated it performance on four different datasets, including 1) GSM8K~\cite{GSM8K}, 2)AQUA~\cite{aqua}, 3) AddSub~\cite{Addsub}, 4) SVAMP~\cite{SVAMP}. Details of datasets are provided in Appendix \ref{sec:dataset}. We conduct experiments using 1 Nvidia 24GB 3090 GPU, with hyperparameters listed in Appendix \ref{sec:Hyperparameters}.

Table~\ref{tab:mathresult} presents the result on 4 arithmetic reasoning benchmarks, demonstrating DenseLoRA’s strong performance with significantly fewer trainable parameters. Notably: at rank=32, DenseLoRA achieves superior performance using only 0.02\% trainable parameters, a $35 \times $ reduction compared to LoRA (0.7\% trainable parameters). At rank=64, DenseLoRA attains 58.5\% accuracy using only 0.06\% trainable parameters, surpassing LoRA's 56.9\% (0.7\% trainable parameters). 
These results further validate DenseLoRA’s effectiveness, demonstrating its parameter efficiency and adaptability across different reasoning tasks.

\begin{figure}[t]
\centering
  \includegraphics[width=\columnwidth]{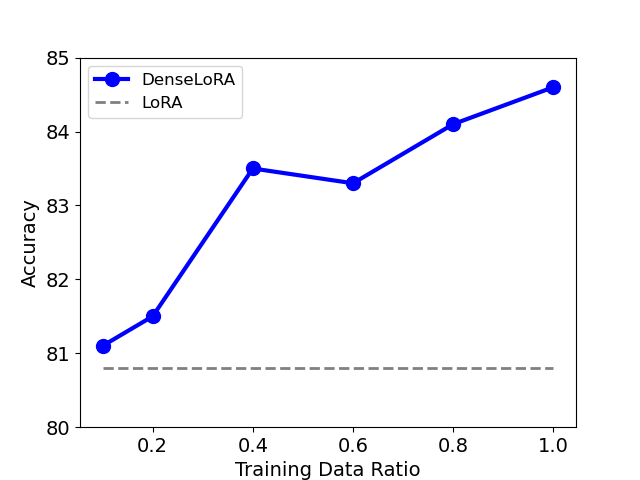}
  \caption{Accuracy(\%) comparison of DenseLoRA and LoRA on the commonsense reasoning 170k dataset with fewer training samples.}
  \label{fig:lowres}
\end{figure}

\subsection{Low Resources Performance} Building on our evaluation of DenseLoRA's parameter efficiency, we now examine its performance under low-resource conditions. We randomly sampled 10\%, 20\%, 40\%, 60\%, and 80\% of the original 170k commonsense reasoning training dataset and repeated the experiments using LLaMA3-8B with a rank of 32. Figure~\ref{fig:lowres} illustrates the relationship between training sample size and performance, with detailed numbers presented in Appendix~\ref{sec:lowresourse}. Notably, DenseLoRA consistently outperforms LoRA across all sample sizes, demonstrating its ability to generalize effectively even with limited data. For instance, DenseLoRA trained with just 10\% of the data achieves an accuracy of 81.1\%, which is 0.3\% higher than LoRA trained with the full dataset. As the number of training samples increases, the performance gap between DenseLoRA and LoRA widens, further highlighting DenseLoRA’s ability to effectively capture complex patterns with reduced data requirements.

\subsection{Tuning Granularity Analysis} This section evaluates the impact of adapting different weight modules using DenseLoRA. Each module is represented by its first letter as follows: (Q)uery, (K)ey, (V)alue, (O)utput, (G)ate, (U)p, (D)own. We conduct experiments using LLaMA3-8B with a rank of 32 on commonsense reasoning training samples. The result, shown Table~\ref{tab:Tuninggra}, highlight several key observations:

Consistent with the original LoRA configuration suggested in~\cite{hu2022lora}, which requires tuning both the Multi-head Attention and MLP layers for optimal performance, DenseLoRA achieves superior accuracy when updating both components.  
Furthermore, adapting the MLP layers proves to be more effective than tuning Multi-head Attention layers. Notably, when DenseLoRA is applied to QKV modules, the model achieves an average of 82.3\%. However, when the UD modules are adapted instead, accuracy improves to 83.8\%, surpassing the QKV configuration. Interestingly, even when only the UD modules are tuned using DenseLoRA (without tuning QKV modules), the accuracy remains at 83.8\%. 
These findings indicate that DenseLoRA is highly efficient than LoRA, achieving superior performance with significantly fewer trainable parameters.

\begin{table}[t]
    \begin{center} 
    \resizebox{\linewidth}{!}{
    \begin{tabular}{cccc}
    \hline 
    \textbf{\#Params(\%)}& \textbf{LoRA} & \textbf{DenseLoRA} & \textbf{Avg.} \\ \hline
    0.70 & QKVUD & -     & 80.8     \\ \hline
    0.25 & QKV   & UD    & 83.8     \\
    0.49 & UD    & QKV   & 83.2     \\
    0.01 & -     & QKV   & 82.3     \\
    0.02 & -     & UD    & 83.8     \\
    0.02 & -     & QKVUD & \textbf{84.6}\\\hline  
    \end{tabular}}
    \end{center}
    \caption{Accuracy(\%) comparison of DenseLoRA with several different tuning granularity fine-tuning LLaMA3-8B. Each module is represented by its first letter as follows: (Q)uery, (K)ey, (V)alue, (O)utput, (U)p, (D)own.}
    \label{tab:Tuninggra}
\end{table}

\begin{table}[t]
    \begin{center} 
    \begin{tabular}{lcccccc}
        \hline 
        \textbf{Method} & \textbf{\# Params(\%)} &\textbf{Avg.} \\ 
       \hline
        DenseLoRA  &  0.02   & \textbf{84.6} \\
        -- Freeze     &  0.03   & 79.5 \\
        -- Only Matrix&  0.02   & 83.3 \\  
        \hline  
    \end{tabular}
    \end{center}
    \caption{Accuracy (\%) comparison of several variants of DenseLoRA.}
    \label{tab:freezeVariant}
\end{table}

\begin{figure}[t]
\centering
  \includegraphics[width=\linewidth]{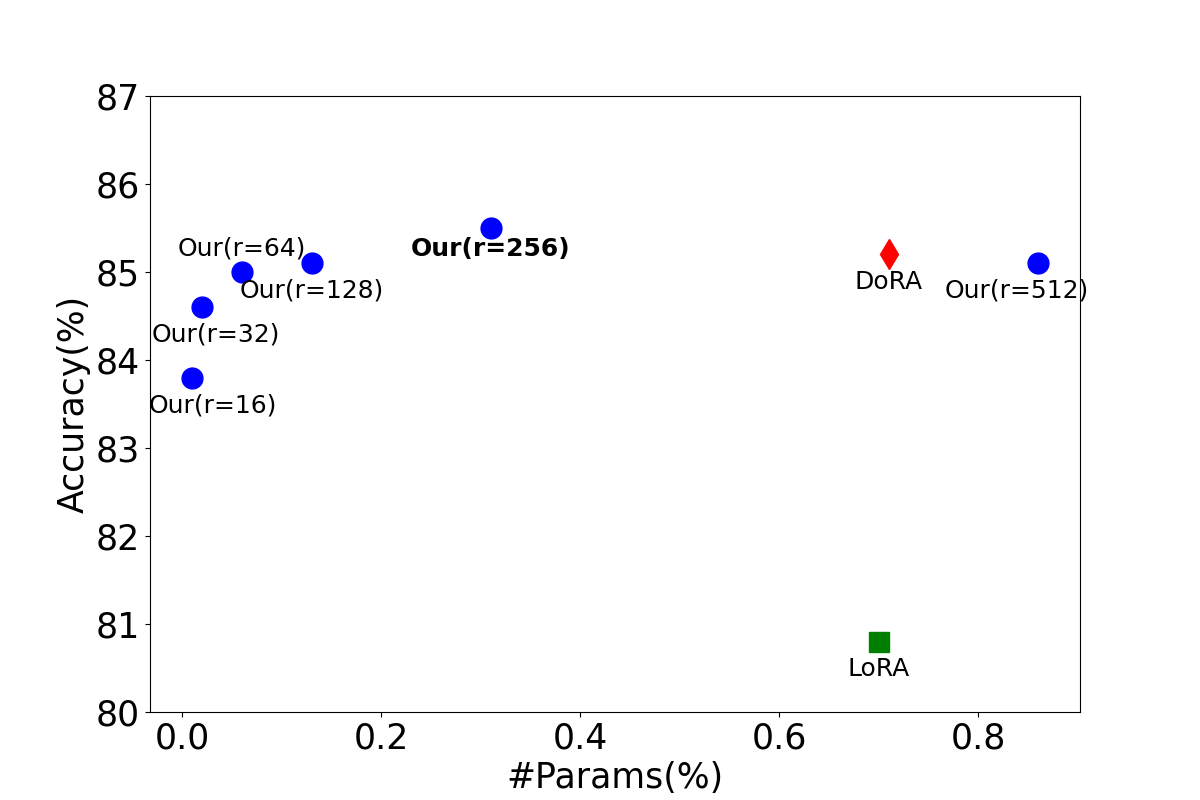} 
  \caption {Evaluate DenseLoRA with rank from 16 to 512.}
    \label{fig:robustness}
\end{figure}

\begin{figure}[t]
\centering
  \includegraphics[width=\linewidth]{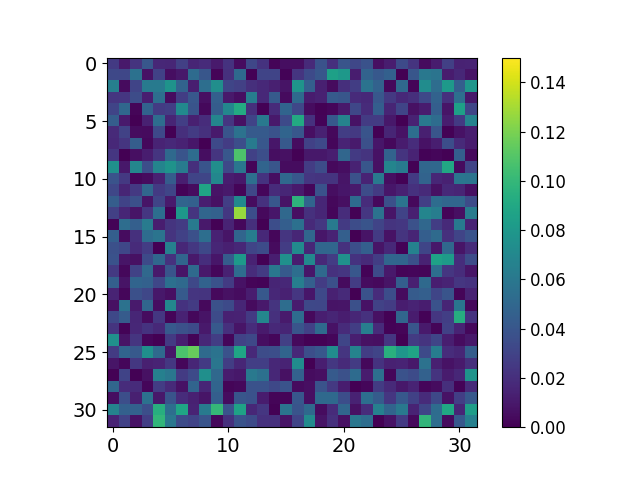} 
  \caption {Increments of matrix $\textbf{\textit{M}}$ during training.}
    \label{fig:deltam}
\end{figure}
\subsection{Robustness of Rank}

We evaluated DenseLoRA with more rank configurations of 128, 256, and 512 on commonsense reasoning tasks using LLaMA3-8B. The corresponding performance is shown in the following Figure~\ref{fig:robustness}. Notably, with a rank of 256 and 0.31\% of the parameters fine-tuned, DenseLoRA outperforms LoRA (80.8\%) and DoRA(85.2\%) with a performance of 85.5\%. Therefore, DenseLoRA performance can be improved by increasing the number of parameters fine-tuned.

\subsection{Understanding the DenseLoRA}
Having established DenseLoRA’s empirical advantages, we now examine its internal mechanisms to better understand the advantages of DenseLoRA. To this end, we conduct a focused investigation on commonsense reasoning tasks using LLaMA3-8B (rank = 32), addressing two key questions: 
1) What is the effectiveness of the representation fine-tuning module that is the $Encoder$ and $Decoder$ modules? 
2) How does the adaptation matrix $\textbf{\textit{M}}$ compare to matrices $A$ and $B$ used in LoRA?

\textbf{Effectiveness of \textit{Encoder} and \textit{Decoder}}: To explore the role of the representation fine-tuning module in DenseLoRA, we evaluate two key variants of the $Encoder$ and $Decoder$ components:
1) \textbf{Freeze}: In this setting, we freeze the parameters of both the $Encoder$ and $Decoder$, keeping them fixed during training, This significantly reduces the number of trainable parameters and is conceptually similar to VeRA~\citep{kopiczko2024vera}. To maintain a comparable number of trainable parameters to DenseLoRA, we compensate by setting the rank to 128. 
2) \textbf{Only Matrix}: This variant isolates the effect of the matrix transformation by removing the activation function, allowing us to assess whether activation functions contribute significantly to representation fine-tuning.

Table~\ref{tab:freezeVariant} presents the experimental result of DenseLoRA compared to its variants.  The results underscore the crucial of the $Encoder$ and $Decoder$ components in the effective fine-tuning of LLMs. Furthermore, the Only Matrix variant, which removes the activation function, shows a performance decrease. This finding indicates that activation functions are not merely auxiliary components but rather play an essential role in enhancing representation fine-tuning.

\textbf{Dense Matrix $\textbf{\textit{M}}$: } To verify the advantages of DenseLoRA's dense matrix representation over LoRA, we compare $\Delta M $ with $\Delta A$ and $\Delta B$ from the same adaptation module and layer during training. Since $A$ and $B$ are significantly larger than $\textbf{\textit{M}}$, we randomly select the slices of $A$ and $B$ that match the size of $\textbf{\textit{M}}$. Figure~\ref{fig:deltam} presents the incremental updates of $\textbf{\textit{M}}$, while a detailed comparison with $A$ and $B$ is provided in Appendix~\ref{sec:comparmab}. From the figure, we observe that $\textbf{\textit{M}}$ exhibits a dense update pattern, where the majority of its parameters actively contribute to the adaptation process. In contrast, $A$ and $B$ show sparse updates, with most of their parameters either remaining unchanged or undergoing minimal modifications.
These findings demonstrate that DenseLoRA is highly parameter-efficient, effectively utilizing a compact adaptation matrix while maintaining strong performance. 

\section{Related Works}
\textbf{Parameter-Efficient Fine-Tuning (PEFT)}: It is a widely adopted strategy aimed at fine-tuning a limited number of parameters in LLMs while keeping the remainder unchanged. These approaches involve fine-tuning only a subset of the existing model parameters or adding new parameters to the model. These methods are designed to reduce the high computational cost of full fine-tuning LLMs~\cite{peftsurvey2023, peft2024survey}. Existing PEFT methods are primarily classified into four types. The first type is known as adapter-based methods. These methods involved inserting adapter layers between the existing layers in LLMs~\cite{adapter, paralleladapter, NEURIPS2021_081be9fd}. The second type involves prompt-based methods~\cite{lester-etal-2021-power, razdaibiedina-etal-2023-residual, li-liang-2021-prefix}. These approaches introduced trainable soft tokens (prompts) into the model's input rather than modifying its internal weights. While these methods involved altering the model’s input or architecture, they resulted in increased inference latency compared to other fine-tuning methods. The third type of method is the low-rank method like LoRA.

\textbf{Low-Rank Adaptation (LoRA)}~\cite{hu2022lora}: It is a technique designed for the low-rank properties of model updates to improve parameter efficiency. Specifically, LoRA used two small matrices to approximate the weight increments during fine-tuning. Recently, there have been numerous methods exploring more efficient LoRA variants. For example, AdaLoRA~\cite{zhang2023adaptive} builds upon LoRA by applying Singular Value Decomposition (SVD) to prune less significant singular values, enhancing update efficiency. Additionally, DoRA~\cite{liu2024dora} decomposes the pre-trained weight into two components: magnitude and direction, allowing LoRA to focus solely on directional updates. Another recent contribution, VeRA~\cite{kopiczko2024vera} introduces the use of “scaling vectors” to adapt frozen random matrices that are shared between layers, significantly reducing the number of trainable parameters compared to traditional LoRA. 

\textbf{Representation Fine-tuning:}
Prior interpretability research has demonstrated that hidden representations in LLMs encode rich semantic information. As a result, representation fine-tuning has emerged as a new type of PEFT method~\cite{wu2024reft, red, yin2024lofit}. For example, RED~\cite{red} modifies the representations at intermediate layers through scaling and biasing operations, significantly reducing the number of trainable parameters. 

Our approach integrates low-rank adaptation with representation fine-tuning, which is different from the existing PEFT method, and achieves superior performance.

\section{Conclusion}
To address weight redundancy in LoRA, we propose Dense Low-Rank Adaptation (DenseLoRA), which significantly reduces the number of trainable parameters while outperforming LoRA. DenseLoRA achieves this by employing a shared $Encoder$ and $Decoder$ across all layers to refine and compress hidden representations before adaptation. Instead of relying on two redundant low-rank matrices, DenseLoRA fine-tunes LLMs using a dense low-rank matrix, leading to more efficient parameter utilization. We evaluate the effectiveness of DenseLoRA on various benchmarks, demonstrating that it achieves performance comparable to LoRA while requiring only 1/70 of the trainable parameters. DenseLoRA presents a more efficient approach to low-rank adaptation while maintaining and even improving model performance compared to existing methods like LoRA.

\section{Limitations}
In this paper, we conduct experiments on commonsense reasoning tasks, and arithmetic reasoning tasks by fine-tuning LLaMA2-7B and LLaMA3-8B. There is a broader range of tasks unexplored using DenseLoRA, such as image generation tasks, and visual instruction tuning tasks. We will apply DenseLoRA in these tasks for future work.

\section{Acknowledgements}
This work is supported by the National Natural Science Foundation of China (No.62206004, No.62272001, No.62072419, No.62406095), and 
the Natural Science Foundation of Anhui Province (No.2308085MF213)


\appendix

\clearpage

\section{Experimental Setting}
\subsection{Dataset} \label{sec:dataset}
\textbf{Commonsense Reasoning:} tasks consist of 8 benchmarks and the details are described as follows:
\begin{itemize}
    \item{BoolQ}~\cite{clark-etal-2019-boolq}: This dataset comprises a collection of yes/no question examples, totaling 15942 examples. These questions are naturally occurring and generated in unprompted and unconstrained settings;
    \item {PIQA}~\cite{bisk2020piqa}: This dataset consists of questions with two solutions that require physical commonsense to answer;
    \item {SIQA}~\cite{sap-etal-2019-social}: This dataset focuses on analyzing people's actions and their social implications;
    \item {HellaSwag}: This dataset consists of commonsense Natural Language Inference (NLI) questions, each featuring a context and multiple endings that complete the context;
    \item {WinoGrande}~\cite{WinoGrande}: This dataset presents a fill-in-a-blank task with binary options. The goal is to select the appropriate option for a given sentence that requires commonsense reasoning;
    \item{ARC-c and ARC-e}~\cite{ARC}: These two datasets are the Challenge Set and Easy Set of ARC~\cite{ARC} dataset, which contains genuine grade-school level, multiple-choice science questions; 
    \item{OBQA}: This dataset comprises questions that require multi-step reasoning, the use of additional common sense knowledge, and thorough text comprehension.
\end{itemize}

\textbf{Arithmetic Reasoning:} tasks consist of 4 benchmarks and the details are described as follows:
\begin{itemize}
    \item {GSM8K}~\cite{GSM8K}: This dataset consists of 8,500 high-quality, linguistically diverse elementary math problems created by humans;
    \item{AQUA}~\cite{aqua}: This dataset comprises 100,000 multiple-choice questions focused on mathematics, encompassing a wide range of topics and varying levels of difficulty;
    \item{SVAMP} (Simple Variations on Arithmetic Math Word Problems)~\cite{SVAMP}: This dataset comprises arithmetic word problems appropriate for fourth-grade students and below;
    \item{AddSub}~\cite{Addsub} This dataset consists of 395 problems specifically focused on addition and subtraction.
\end{itemize}

\begin{table}[t]
  \centering
    \setlength{\tabcolsep}{3pt}
    \resizebox{\linewidth}{!}{
  \begin{tabular}{ccccc}
    \hline
    \textbf{HyperParmaters}    & \multicolumn{2}{c}{\textbf{LLaMA2-7B}} & \multicolumn{2}{c}{\textbf{LLaMA3-8B}}  \\
    \hline
    Rank r       &  16  &  32  & 16   & 32             \\
    $\alpha$     &  32  &  64  & 32   & 64             \\
    Dropout      &  \multicolumn{4}{c}{0.05}           \\
    Optimizer    &     \multicolumn{4}{c}{AdamW}       \\
    LR           &   \multicolumn{4}{c}{3e-4}          \\
    LR Scheduler &     \multicolumn{4}{c}{Linear}      \\
    Batch Size   &       \multicolumn{4}{c}{16}         \\
    Warmup Stemps&     \multicolumn{4}{c}{100}          \\
    Epochs       &      \multicolumn{4}{c}{2}           \\
    Where        & \multicolumn{4}{c}{Q, K, V, Up, Down} \\
    \hline
  \end{tabular}}
  \caption{\label{tab:Hyperparameters}
    The hyperparameters for DenseLoRA on the commonsense reasoning tasks.
  }
\end{table}

\begin{table}[t]
  \centering
  \begin{tabular}{ccccc}
    \hline
    \textbf{HyperParmaters}  & \multicolumn{2}{c}{\textbf{LLaMA3-8B}}  \\
    \hline
    Rank r       &  16  &  32           \\
    $\alpha$     &  32  &  64             \\
    Dropout      &  \multicolumn{2}{c}{0.05}           \\
    Optimizer    &     \multicolumn{2}{c}{AdamW}       \\
    LR           &   \multicolumn{2}{c}{3e-4}          \\
    LR Scheduler &     \multicolumn{2}{c}{Linear}      \\
    Batch Size   &       \multicolumn{2}{c}{16}         \\
    Warmup Stemps&     \multicolumn{2}{c}{100}          \\
    Epochs       &      \multicolumn{2}{c}{2}           \\
    Where        & \multicolumn{2}{c}{Q, K, V, Up, Down} \\
    \hline
  \end{tabular}
  \caption{\label{tab:Hyperparametersari}
    The hyperparameters for DenseLoRA on the arithmetic reasoning tasks.
  }
\end{table}

\subsection{Hyperparameters} \label{sec:Hyperparameters}
Table~\ref{tab:Hyperparameters} shows the detailed hyperparameters for commonsense reasoning tasking when fine-tuning the LLaMA3-8B and LLaMA2-7B. Table~\ref{tab:Hyperparametersari} shows the detailed hyperparameters for arithmetic reasoning tasking when fine-tuning the LLaMA3-8B. During the inference, we set the hyperparameters $max\_new\_tokens=128$.

\section{Additional Experimental}

\subsection{Low Resources Experimental}\label{sec:lowresourse}
Table~\ref{tab:lowresourseresults} shows the details for low resources setting of commonsense reasoning tasks when fine-tuning the LLaMA3-8B and using the rank equal to 32.
\begin{table*}[ht]
    \begin{center} 
    \renewcommand{\arraystretch}{1.1}
    \setlength{\tabcolsep}{3pt}
    \begin{tabular}{l|cccccccc|c}
        \hline 
        \textbf{Ratio} & \textbf{BoolQ} & \textbf{PIQA} &  \textbf{SIQA} & \textbf{HellaS.} & \textbf{WinoG.} & \textbf{ARC-e} & \textbf{ARC-c} &  \textbf{OBQA} & \textbf{Avg.}\\ \hline
  10\% & 71.6 & 85.0 & 76.4 & 90.3 & 79.2 & 89.6 & 77.7 & 78.6 & 81.1 \\
  20\% & 63.5 & 86.6 & 78.6 & 92.2 & 81.8 & 77.7 & 89.8 & 81.8 & 81.5 \\ 
  40\% & 73.4 & 86.8 & 78.7 & 93.2 & 85.1 & 77.5 & 90.0 & 83.0 & 83.6 \\ 
  60\% & 68.6	& 87.4 & 79.5 & 93.9 & 84.4	& 78.2 & 90.2 & 84.4 & 83.3 \\
  80\% & 72.9	& 88.4 & 79.3 &	94.1 & 85.5 & 78.2 & 90.6 &	84.0 & 84.1 \\
        \hline  
    \end{tabular}
    \end{center}
    \caption{\label{tab:lowresourseresults}Accuracy(\%) comparison of fewer training samples methods fine-tuning LLaMA3-8B on the commonsense reasoning 170k dataset in~\cite{hu-etal-2023-llm}. \textbf{Ratio} denotes the ratio of trained parameters. We set the rank = 32.}
\end{table*}

\subsection{Tuning Granularity Analysis} \label{sec:tungran}
This section is a detailed experimental result of adapting different weight modules using DenseLoRA. Each module is represented by its first letter as follows: (Q)uery, (K)ey, (V)alue, (O)utput, (G)ate, (U)p, (D)own. We conduct experiments using LLaMA3-8B with a rank of 32 on commonsense reasoning training samples. The result, shown Table~\ref{tab:detailgran}

\begin{table*}[ht]
    \label{tab:tungra results}
    \begin{center} 
    \renewcommand{\arraystretch}{1.1}
    \setlength{\tabcolsep}{3pt}
    \resizebox{\linewidth}{!}{
    \begin{tabular}{cccccccccccc}
        \hline 
        \multirow{2}*{\textbf{\# Params(\%)}} & \multicolumn{2}{c}{\textbf{Method}} & \multirow{2}*{\textbf{BoolQ}} & \multirow{2}*{\textbf{PIQA}} &  \multirow{2}*{\textbf{SIQA}} & \multirow{2}*{\textbf{HellaS.}} & \multirow{2}*{\textbf{WinoG.}} & \multirow{2}*{\textbf{ARC-e}} & \multirow{2}*{\textbf{ARC-c}} &  \multirow{2}*{\textbf{OBQA}} & \multirow{2}*{\textbf{Avg.}}\\ 
        & \textbf{LoRA} & \textbf{DenseLoRA}\\
        \hline
  0.25 & QKV & UD  & 74.3	& 86.7 & 79.9 & 94.2 & 85.6 & 77.0 & 88.5 & 83.8 & 83.8 \\
  0.49 & UD  & QKV & 72.5	& 88.0 & 81.4 & 93.8 & 84.1 & 74.7 & 86.8 & 84.0 & 83.2 \\ 
  0.01 & -   & QKV & 72.2	& 86.5 & 78.6 & 90.8 & 82.4	& 77.0 & 90.2 & 80.4 & 82.3 \\ 
  0.02 & -   & UD  & 71.0 & 87.3 & 78.9 & 94.0 & 84.8	& 79.3 & 90.3 & 84.4 & 83.8 \\
        \hline 

    \end{tabular}}
    \end{center}
    \caption{Accuracy(\%) comparison of several different tuning granularity of DenseLoRA fine-tuning LLaMA3-8B. Each module is represented by its first letter as follows: (Q)uery, (K)ey, (V)alue, (O)utput, (U)p, (D)own.}
    \label{tab:detailgran}
\end{table*}
\subsection{Compared M and A, B} \label{sec:comparmab}
\begin{figure*}[ht]
    \begin{minipage}[t]{0.33\textwidth}
		\centering
		\includegraphics[width=\textwidth]{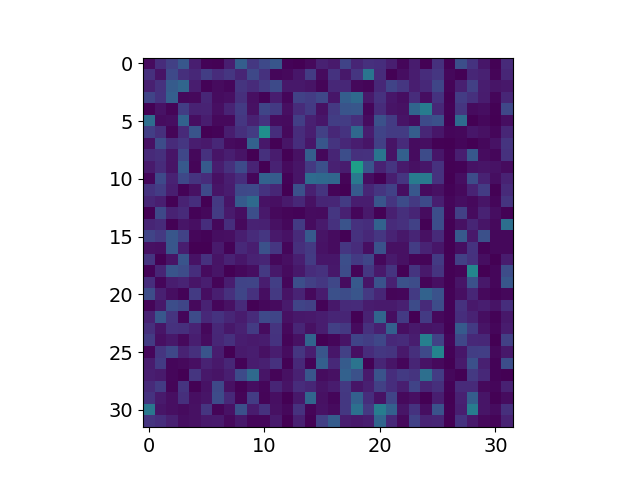}
		\subcaption{(a) $\Delta A$} 
        \label{fig:matrixa}
    \end{minipage}
    \begin{minipage}[t]{0.33\textwidth}
		\centering
		\includegraphics[width=\textwidth]{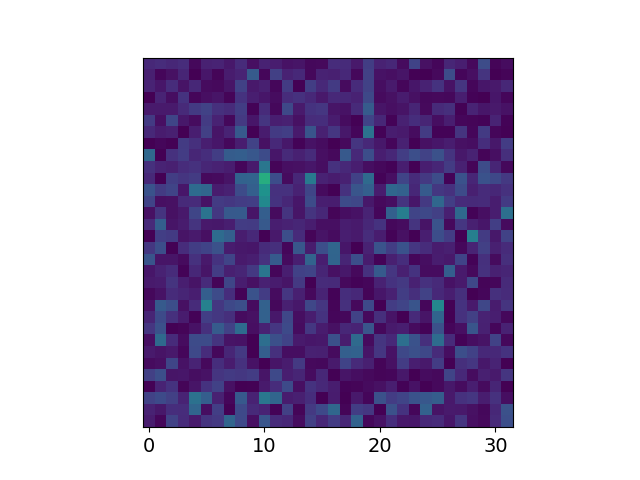}
		\subcaption{(b) $\Delta B$}
        \label{fig:matrixb}
	\end{minipage}
    \begin{minipage}[t]{0.33\textwidth}
		\centering
		\includegraphics[width=\textwidth]{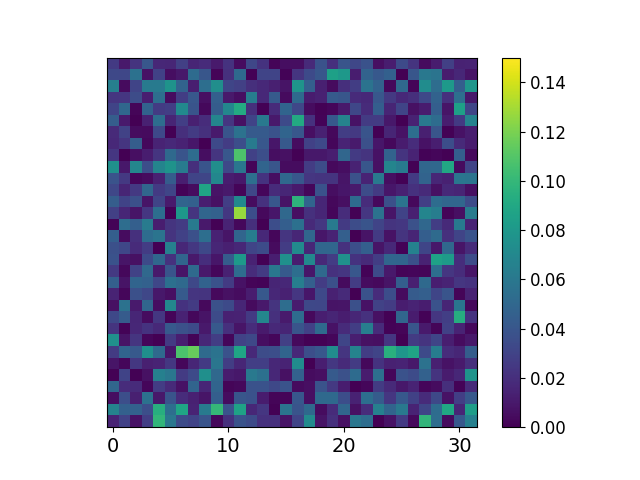}
		\subcaption{(c) $\Delta \textbf{\textit{M}}$}
        \label{fig:matrixm}
	\end{minipage}
  \caption{Increamts of matrices $A$ and $B$ of LoRA compared to matrix \textbf{\textit{M}} of DenseLoRA. We randomly select the slices of a
small set of $A$ and $B$ for demonstration.}
  \label{fig:delta}
\end{figure*}

we compare $\Delta M $ with $\Delta A$ and $\Delta B$ from the same adaptation module and layer during training. Since $A$ and $B$ are significantly larger than $\textbf{\textit{M}}$, we randomly select the slices of $A$ and $B$ that match the size of $\textbf{\textit{M}}$. The detailed result is shown in Figure~\ref{fig:delta}.

\end{document}